\documentclass[runningheads]{llncs}
\usepackage{graphicx}
\usepackage{tikz}
\usepackage{comment}
\usepackage{amsmath,amssymb} 
\usepackage{color}
\usepackage{float}
\usepackage{epsfig}
\usepackage{bm}
\usepackage{multirow}
\usepackage{subfigure}
\usepackage{threeparttable}
\usepackage{booktabs}
\usepackage[OT1]{fontenc} 
\usepackage{color}
\begin{document}
%
\title{PathoTune: Adapting Visual Foundation Model to Pathological Specialists}
%
%

\author{Jiaxuan Lu\inst{1} \and
Fang Yan\inst{1} \and
Xiaofan Zhang\inst{1,3} \and
Yue Gao\inst{2} \and
Shaoting Zhang\inst{1,}\thanks{S. Zhang is the corresponding author. \\ This work was supported by Shanghai Artificial Intelligence Laboratory.}
}

%
\authorrunning{J. Lu et al.}
%

\institute{Shanghai Artificial Intelligence Laboratory, Shanghai 200232, China \and
Tsinghua University, Beijing 100084, China \and
Shanghai Jiao Tong University, Shanghai 200240, China
\email{\{lujiaxuan,yanfang,zhangshaoting\}@pjlab.org.cn} 
}

\maketitle              
%
\begin{abstract}
As natural image understanding moves towards the pretrain-finetune era, research in pathology imaging is concurrently evolving. Despite the predominant focus on pretraining pathological foundation models, how to adapt foundation models to downstream tasks is little explored. For downstream adaptation, we propose the existence of two domain gaps, \textit{i.e.}, the Foundation-Task Gap and the Task-Instance Gap. To mitigate these gaps, we introduce \textbf{PathoTune}, a framework designed to efficiently adapt pathological or even visual foundation models to pathology-specific tasks via multi-modal prompt tuning. The proposed framework leverages Task-specific Visual Prompts and Task-specific Textual Prompts to identify task-relevant features, along with Instance-specific Visual Prompts for encoding single pathological image features. Results across multiple datasets at both patch-level and WSI-level demonstrate its superior performance over single-modality prompt tuning approaches. Significantly, PathoTune facilitates the direct adaptation of natural visual foundation models to pathological tasks, drastically outperforming pathological foundation models with simple linear probing. The code is available at \url{https://github.com/openmedlab/PathoDuet}.

\end{abstract}
\section{Introduction}
Pathological image diagnostics stands as a critical step that informs clinical decisions by examining and interpreting stained images at the cellular level. Computational pathology integrates machine learning techniques that promise to revolutionize the approach to disease detection and analysis. In recent years, many deep learning-based pathology diagnostic methods have been explored, which can be categorized into patch-level~\cite{xu2017large,xu2015deep,kallen2016towards} and WSI-level~\cite{wang2019weakly,xu2017parallel,pal2022bag} frameworks. However, these models need to be individually trained for specific downstream tasks, \textit{e.g.}, training a separate model to recognize breast cancer or Gleason grade of prostate, lacking in flexibility and generality.

As the field of language processing and image analysis has transitioned into the pretrain-finetune era, computational pathology has also entered the paradigm of foundation models and efficient finetuning, where it is desirable to train a generalized foundation model that can solve all downstream tasks. Recent advances have explored how self-supervised pretraining on large datasets can be used to develop pathological foundation models, including CTransPath~\cite{wang2022transformer}, HIPT~\cite{chen2022scaling}, Pathoduet~\cite{hua2023pathoduet}, Virchow~\cite{vorontsov2023virchow}, \textit{etc.}
Despite these advancements, it is demonstrated that even with pathological foundation models, satisfactory performance cannot be achieved without finetuning~\cite{sikaroudi2023generalization}. While some works in the field of natural image analysis have explored the Parameter-Efficient Fine-Tuning (PEFT)~\cite{chen2022adaptformer,liu2022p}, how to effectively transfer the foundation models to downstream pathological tasks has received little attention.

\begin{figure*}[t]
  \centering
  \includegraphics[width=\textwidth]{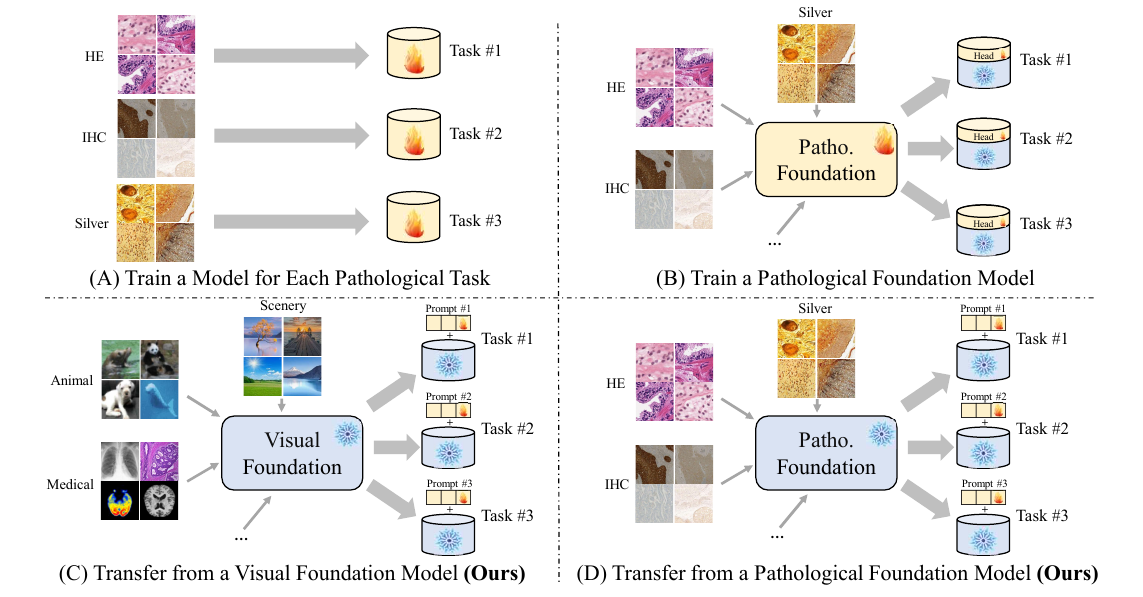}
  \caption{Compared to traditional paradigms of training separate models for each task or training a pathological foundation model, PathoTune directly adapts a visual or pathological foundation model to downstream tasks using multi-modal prompts.}
  \label{Fig:first_page_mode_compare}
\end{figure*}

This paper argues for the importance of efficiently adapting a generalist foundation model to downstream specialized models for pathological tasks. We propose that there exist two primary domain gaps in this process: the Foundation-Task Gap (FTG) and the Task-Instance Gap (TIG). FTG refers to the domain difference between the data encountered by the foundation model and the downstream pathological task or dataset, while TIG denotes the discrepancy between a specific image and the average distribution of images in the dataset, \textit{e.g.}, varied staining variations inherent to each pathological image.

To address these challenges, we introduce \textbf{PathoTune}, a framework that employs multi-modal prompt tuning to adapt a foundation model for pathology-specific tasks with a minor parameter increment. The foundation model can be either a pretrained natural visual model or a pathological foundation model, as shown in Fig.~\ref{Fig:first_page_mode_compare}. PathoTune leverages Task-specific Visual Prompts (TVP) and Task-specific Textual Prompts (TTP) to bridge the FTG by encoding task-related information. Additionally, it utilizes a Visual Refine Module to generate Instance-specific Visual Prompts (IVP) for addressing the TIG. 
The results from multiple datasets at both patch-level and WSI-level demonstrate that PathoTune not only outperforms state-of-the-art (SOTA) PEFT methods relying on single-modal prompts but also significantly exceeds elaborately pretrained pathological foundation models with linear probing.

%
%

\section{Related Work}
\textbf{Traditional Pathology Modeling.} In digital pathology, identifying cancer in Whole Slide Images (WSIs) poses a significant challenge due to their large size. Xu \textit{et al.}~\cite{xu2015deep} leverages CNNs pretrained on ImageNet to extract features from WSI patches for classification. To diagnose with WSI-level labels, Multi-Instance Learning (MIL) is utilized in a series of works~\cite{wang2019weakly,xu2017parallel,courtiol2018classification}, integrating CNNs with MIL for WSI classification. 
Additionally, Transformer models are being investigated for a more integrated WSI analysis by feeding features from numerous patches~\cite{shao2021transmil,li2021dt}. Regardless of the backbone network, these methods require either training from scratch or full finetuning on a pretrained model, with a separate model needed for each specific task.

\textbf{Pathological Foundation Model.} With the emergence of foundation models in natural language processing~\cite{vaswani2017attention,brown2020language}, computer vision~\cite{he2020momentum,he2022masked}, \textit{etc.}, recent studies have explored the development of pathological foundation models based on self-supervised learning. Studies like Huang \textit{et al.}~\cite{huang2021integration} and Ciga \textit{et al.}~\cite{ciga2022self} apply contrastive learning to pathological patches. CTransPath~\cite{wang2022transformer} enhances the MoCo v3 framework with a pseudo positive selection mechanism to improve similarity handling between patches. 
Pathoduet~\cite{hua2023pathoduet} builds on MoCo v3 with additional pretraining tasks for cross-scale and cross-stain challenges. Large-scale data utilization includes HIPT's~\cite{chen2022scaling} hierarchical pyramid ViT pretraining on 10,678 WSI slides, UNI~\cite{chen2023general} employing 100,000 slides with the DINO v2 framework, and Virchow~\cite{vorontsov2023virchow} using 1.5 million slides.

\textbf{Parameter-Efficient Fine-Tuning.} PEFT has become a prominent and efficient alternative in natural language processing~\cite{ben-zaken-etal-2022-bitfit,liu2022p}, offering accuracy comparable to full finetuning but with fewer parameters and reduced storage. In computer vision, Adaptformer~\cite{chen2022adaptformer} finetunes visual adapters in foundation models for diverse tasks.
Prompt tuning~\cite{jia2022visual,tu2023visual} emerges as an alternative, enabling task transfer without altering the network's structure. Jia~\textit{et al.} introduces VPT~\cite{jia2022visual} with learnable tokens as visual prompts, and Sohn~\textit{et al.}~\cite{sohn2023visual} proposes generating prompt tokens using a generator. VQT~\cite{tu2023visual} utilizes ``query''-only learnable tokens for further parameter reduction. 
How to adapt the foundation model to pathological downstream tasks is worth exploring as well.

\begin{figure*}[t]
  \centering
  \includegraphics[width=1.0\linewidth]{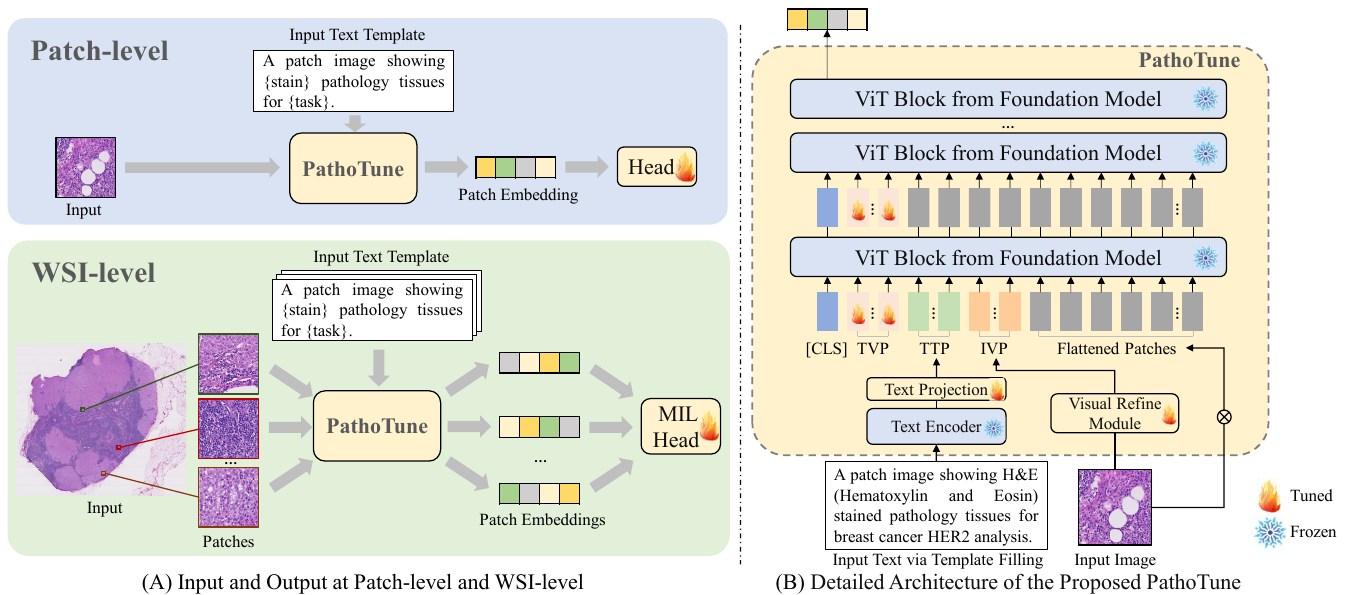}
  \caption{Overview of the proposed PathoTune. (A) The input and output of PathoTune for both patch-level and WSI-level tasks. (B) Detailed architecture of PathoTune, encompassing the Task-specific Visual Prompts (TVP), Task-specific Textual Prompts (TTP) and Instance-specific Visual Prompts (IVP).}
  \label{Fig:pipeline}
\end{figure*}

\section{Methodology}
\subsection{Problem Formulation\label{sec:3.1}}
Assuming that the data distribution of the natural image and the pathology image is represented as $F$ and $D$, respectively, the corresponding visual foundation model and the ideal pathology model are represented as $\Phi(\cdot)$ and $\Psi(\cdot)$. In the adaptation of foundation models to downstream tasks, we identify the existence of two domain gaps: the Foundation-Task Gap and the Task-Instance Gap. 

\begin{itemize}
\item Foundation-Task Gap (FTG): The gap between the data domain $F$ pretrained by the foundation model $\Phi(\cdot)$ and downstream pathological domain $D$, which is relevant to the specific task.
\item Task-Instance Gap (TIG): Domain gap between each instance image in the task-specific dataset and the dataset's average data distribution, including nuances such as staining and glandular structure variations.
\end{itemize}

In downstream adaptation, the FTG not only encompasses the discrepancy between the natural image domain $F$ and the pathology domain $D$, but also reflects the significant divergence between the visual foundation model $\Phi(F)$ and the ideal pathology model $\Psi(D)$. To bridge this gap, we propose the employment of task-specific prompts $P_{task}$, which are designed to adapt the visual foundation model $\Phi(F)$ to the pathology domain $D$, denoted as $\Phi(D;P_{task})$. The proposed task-specific prompts $P_{task}$ are aimed at minimizing
\begin{equation}
    \min_{P} \left\| \Phi(D; P_{task}) - \Psi(D) \right\|.
\end{equation}
Particularly, the $\Phi(\cdot)$ can be either a visual or a pathological foundation model.

For each individual instance image $x \in D$ within a pathological dataset, the TIG is quantified as the variance $\sigma(D)$, representing the dispersion of the dataset's distribution. To depict the specificity of the embedding of an instance image $\Phi(x;P_{task})$ compared to the mean value $\Phi(\overline{x};P_{task})$, we introduce the instance-specific prompts $P_{ins}$ which are designed to minimize
\begin{equation}
    \min_{P} \mathbb{E}_{x\in D} \left[ \left\| \Phi(x; P_{ins}) - \Psi(x) \right\| \right],
\end{equation}
where $\Psi(x)$ denotes the embedding derived from feeding image $x$ into the ideal pathology model.

\subsection{Multi-Modal Prompt Design}
In response to the two domain gaps inherent in downstream pathological adaptation, the proposed PathoTune introduces three kinds of prompts, including Task-specific Visual Prompts (TVP), Task-specific Textual Prompts (TTP), and Instance-specific Visual Prompts (IVP). The TVP and TTP are designed as task-specific prompts, with the purpose of relieving the Foundation-Task Gap. Conversely, the IVP serves as the instance-specific prompts. From a modal perspective, both TVP and IVP are categorized as visual prompts, while TTP operates as textual prompts. The complete pipeline as well as the inputs and outputs of PathoTune are shown in Fig.~\ref{Fig:pipeline}.

\textbf{Task-specific Visual Prompts.} To mitigate the Foundation-Task Gap, it is crucial to convey task-specific information to the foundation model. In this context, we interpret the visual prompt explored in existing works~\cite{jia2022visual,tu2023visual} as a type of ``soft'' prompt with learnable task-specific information relevant to the downstream pathological domain. Specifically, the Task-specific Visual Prompts (TVP) introduces several learnable tokens into each layer of the Vision Transformer (ViT). For the ViT with $L$ layers, let $P_{TVP}^l \in \mathbb{R}^{N \times C}$ be the matrix of learnable tokens at layer $l$, where $N$ is the number of tokens and $C$ is the token dimension. The corresponding $N$ TVP tokens $P_{TVP}^l$ are prepended to the patch embedding $E^l$ before being fed into the $l$-th layer.

\textbf{Task-specific Textual Prompts.} In addition to the ``soft'' visual prompts that promote token self-learning, we consider textual descriptions as another approach profiling the downstream pathological task and dataset. The proposed Task-specific Textual Prompts (TTP) $P_{TTP}\in \mathbb{R}^{T \times C}$ utilizes a text template filled with specific stain (\textit{e.g.}, HE or IHC) and task information to generate text embeddings, which are then aligned with other tokens through a frozen text encoder $\theta^{TE}$ and a tunable text projection layer $\theta^{TP}$. Assuming the text template be $P_{text}$ = ``A patch image showing \{stain\} pathology tissues for \{task\}''. The text embedding for a given task is obtained as
\begin{equation}
P_{TTP} = f_{TP} \left( f_{TE} (P_{text}; \theta_{TE}); \theta_{TP} \right),
\end{equation}
where $\theta_{TE}$ and $\theta_{TP}$ are the parameters of the text encoder and the text projection layer, respectively. The text projection layer not only formalizes the feature dimension, but also aligns the text features with flattened patches and prompts.

\textbf{Instance-specific Visual Prompts.} The Instance-specific Visual Prompts (IVP) $P_{IVP}\in \mathbb{R}^{M \times C}$ targets the Task-Instance Gap (TIG) by capturing unique characteristics of individual pathological instances. Specifically, we propose a lightweight Visual Refine Module (VRM) $f_{VRM}$ to extract the specific staining and glandular features relative to a single patch image. For a given instance image $x \in D$, the VRM processes it into a coarse-grained embedding, which is then replicated into $M$ tokens, expressed as
\begin{equation}
P_{IVP} = f_{VRM}(x; \theta_{VRM}),
\end{equation}
where $\theta_{VRM}$ denotes the tunable parameters within the VRM. Different from the TVP and TTP which have fixed tokens for a specific dataset, the IVP is instance-wise, with tokens generated for each input image.

\subsection{Overall Procedure}

The proposed PathoTune appends the above three types of prompts to the input of the ViT structure derived from the foundation model. Assuming that the pathological image is flattened into $K$ tokens, the embedding of the $l$-th layer is expressed as $E^{l-1} \in \mathbb{R}^{K \times C}$. Thus, the first layer of ViT can be represented as
\begin{equation}
[V^1, E^1] = \Phi^1 \left( [V^0, P^0_{TVP}, P_{TTP}, P_{IVP}, E^0] \right),
\end{equation}
where $V^l$ is the [CLS] token, and $\Phi^l(\cdot)$ denotes the Transformer layer of the $l$-th layer. Unlike the first layer which requires three types of tokens, subsequent layers only need to replace the TVP tokens using $P_{TVP}^{l-1}$, expressed as
\begin{equation}
[V^l, E^l] = \Phi^l \left( [V^{l-1}, P^{l-1}_{TVP}, E^{l-1}] \right),
\end{equation}
where the last layer of the [CLS] token $V^L$ is fed into the tunable patch-level or WSI-level head to classify pathological images.
With the proposed prompts, we can reuse the extensive knowledge embedded in the foundation model, requiring only finetuning the prompts and the head for adaptation with a far lesser number of trainable parameters. Additionally, compared to previous paradigms (Fig.~\ref{Fig:first_page_mode_compare} (A) and (B)) required to train a specialized pathology model $\Psi(\cdot)$ based on the pathological dataset $D$, the proposed paradigm offers the support for a multitude of tasks through a shared foundation model augmented by specialized prompts.

\section{Experiments and Results}
\textbf{Datasets.} We conduct a comprehensive evaluation of PathoTune across extensive pathology datasets, including both public datasets, \textit{i.e.}, BCI~\cite{liu2022bci}, NCT~\cite{kather2018100}, SICAPv2~\cite{silva2020going}, and the private RJ-Prost dataset. These datasets span patch-level (BCI, NCT) and WSI-level (SICAPv2, RJ-Prost) tasks, covering various organs and staining types (HE and IHC). Among them, RJ-Prost is a proprietary dataset from an anonymous hospital focusing on prostate Gleason grading, which includes 1,042 WSIs and four categories: negative, grade 3, grade 4, and grade 5. The BCI dataset which contains both HE and IHC stains is divided into ``BCI-HE'' and ``BCI-IHC'' for specific experiments. For dataset division, BCI adheres to the official guideline, with identical validation and test sets. NCT follows the protocol established by Bian \textit{et al}~\cite{bian2022multiple}. For remaining datasets without standardized division protocols, we allocate data into training, validation, and test sets in a 7:2:1 ratio. 4-fold cross-validation is applied to all except BCI.

\textbf{Implementations.} We transfer both visual foundation models (ViT pretrained on ImageNet) and pathological foundation models (HIPT~\cite{chen2022scaling} and Pathoduet~\cite{hua2023pathoduet}) to each downstream dataset individually. HIPT employs ViT-S, while Pathoduet utilizes ViT-B as the backbones for comparisons. The text encoder in TTP leverages the pretrained BERT~\cite{devlin2018bert}, while the VRM module in TVP initializes with the first 4 layers of ResNet18. In most experiments, the token number for TVP, TTP, and IVP is set at 10, 2, and 2, respectively, with a batch size of 32. We employ the RAdam~\cite{liu2019variance} optimizer at a learning rate of 0.0002.

\begin{table*}[!t]
  \renewcommand\arraystretch{0.7}
  \begin{center}
  \caption{Ablation results (\%) based on different foundation models on multiple datasets, where ``FT'' stands for full finetuning and ``LP'' stands for linear probing.}
  \label{tab:dataset_comparison}
  \setlength{\tabcolsep}{0.35 mm}{
\begin{tabular}{c|c|ccc|cccccc|cccc}
\hline
& & \multicolumn{3}{c|}{} & \multicolumn{6}{c|}{Patch-level} & \multicolumn{4}{c}{WSI-level}\\ \cline{6-15}
& & \multicolumn{3}{c|}{\multirow{-2}{*}{Prompts}}& \multicolumn{2}{c|}{BCI-HE}& \multicolumn{2}{c|}{BCI-IHC} & \multicolumn{2}{c|}{NCT}& \multicolumn{2}{c|}{SICAPv2} & \multicolumn{2}{c}{RJ-Prost}\\
\multirow{-3}{*}{Found.}& \multirow{-3}{*}{Mode}& TTP & TVP & IVP & AUC& \multicolumn{1}{c|}{F1} & AUC& \multicolumn{1}{c|}{F1} & AUC& F1& AUC& \multicolumn{1}{c|}{F1} & AUC& F1 \\ \hline
& {\color[HTML]{C0C0C0} FT} & {\color[HTML]{C0C0C0} } & {\color[HTML]{C0C0C0} } & {\color[HTML]{C0C0C0} } & {\color[HTML]{C0C0C0} 94.2} & \multicolumn{1}{c|}{{\color[HTML]{C0C0C0} 76.0}} & {\color[HTML]{C0C0C0} 97.0} & \multicolumn{1}{c|}{{\color[HTML]{C0C0C0} 85.4}} & {\color[HTML]{C0C0C0} 99.8} & {\color[HTML]{C0C0C0} 95.3}& {\color[HTML]{C0C0C0} 94.2} & \multicolumn{1}{c|}{{\color[HTML]{C0C0C0} 74.9}} & {\color[HTML]{C0C0C0} 96.8} & {\color[HTML]{C0C0C0} 77.2} \\
& LP& & & & 54.4& \multicolumn{1}{c|}{15.7}& 63.7& \multicolumn{1}{c|}{26.3}& 97.4& 84.5 & 82.9& \multicolumn{1}{c|}{11.8}& 91.7& 51.1 \\
& &\checkmark & & & 63.1& \multicolumn{1}{c|}{16.2}& 68.0& \multicolumn{1}{c|}{30.1}& 98.5& 89.9 & 84.5& \multicolumn{1}{c|}{25.0}& 93.1& 55.6\\
& & &\checkmark & & 85.8& \multicolumn{1}{c|}{67.5}& 72.5& \multicolumn{1}{c|}{31.1}& 99.2& 91.5 & 87.4& \multicolumn{1}{c|}{59.0}& 93.7& 60.9\\
& & & &\checkmark & 92.5& \multicolumn{1}{c|}{75.5}& 96.0& \multicolumn{1}{c|}{81.5}& 99.4& 92.1 & 91.3& \multicolumn{1}{c|}{68.3}& 95.2& 74.0\\
\multirow{-6}{*}{\begin{tabular}[c]{@{}c@{}}ImageNet\\ (ViT-S)\end{tabular}}& \multirow{-4}{*}{Ours}&\checkmark &\checkmark &\checkmark & \textbf{93.2} & \multicolumn{1}{c|}{\textbf{76.1}} & \textbf{97.3} & \multicolumn{1}{c|}{\textbf{84.3}} & \textbf{99.7} & \textbf{92.4}& \textbf{94.3} & \multicolumn{1}{c|}{\textbf{74.8}} & \textbf{96.8} & \textbf{76.4} \\ \hline
& {\color[HTML]{C0C0C0} FT} & {\color[HTML]{C0C0C0} } & {\color[HTML]{C0C0C0} } & {\color[HTML]{C0C0C0} } & {\color[HTML]{C0C0C0} 94.7} & \multicolumn{1}{c|}{{\color[HTML]{C0C0C0} 79.9}} & {\color[HTML]{C0C0C0} 95.3} & \multicolumn{1}{c|}{{\color[HTML]{C0C0C0} 82.1}} & {\color[HTML]{C0C0C0} 99.8} & {\color[HTML]{C0C0C0} 94.2}& {\color[HTML]{C0C0C0} 95.5} & \multicolumn{1}{c|}{{\color[HTML]{C0C0C0} 79.9}} & {\color[HTML]{C0C0C0} 97.3} & {\color[HTML]{C0C0C0} 80.7} \\
& LP& & & & 53.1& \multicolumn{1}{c|}{14.5}& 61.5& \multicolumn{1}{c|}{29.0}& 98.4& 84.9 & 86.5& \multicolumn{1}{c|}{46.9}& 92.5& 54.7\\ 
& &\checkmark & & & 58.2& \multicolumn{1}{c|}{15.8}& 65.5& \multicolumn{1}{c|}{31.0}& 98.9& 90.2 & 87.3& \multicolumn{1}{c|}{55.3}& 93.5& 58.8\\
& & &\checkmark & & 59.2& \multicolumn{1}{c|}{15.7}& 67.9& \multicolumn{1}{c|}{32.2}& 99.1& 91.3 & 89.8& \multicolumn{1}{c|}{64.9}& 94.1& 67.4\\
& & & &\checkmark & 92.8& \multicolumn{1}{c|}{72.7}& 90.1& \multicolumn{1}{c|}{65.1}& 99.6& 92.4 & 92.9& \multicolumn{1}{c|}{72.5}& 96.3& 74.8\\
\multirow{-6}{*}{\begin{tabular}[c]{@{}c@{}}HIPT\\ (ViT-S)\end{tabular}}& \multirow{-4}{*}{Ours}&\checkmark &\checkmark &\checkmark & \textbf{93.4} & \multicolumn{1}{c|}{\textbf{75.4}} & \textbf{96.8} & \multicolumn{1}{c|}{\textbf{82.8}} & \textbf{99.8} & \textbf{94.0}& \textbf{95.4} & \multicolumn{1}{c|}{\textbf{79.3}} & \textbf{97.0} & \textbf{80.5} \\ \hline
& {\color[HTML]{C0C0C0} FT} & {\color[HTML]{C0C0C0} } & {\color[HTML]{C0C0C0} } & {\color[HTML]{C0C0C0} } & {\color[HTML]{C0C0C0} 95.1} & \multicolumn{1}{c|}{{\color[HTML]{C0C0C0} 81.4}} & {\color[HTML]{C0C0C0} 97.5} & \multicolumn{1}{c|}{{\color[HTML]{C0C0C0} 86.4}} & {\color[HTML]{C0C0C0} 99.7} & {\color[HTML]{C0C0C0} 95.2}& {\color[HTML]{C0C0C0} 97.2} & \multicolumn{1}{c|}{{\color[HTML]{C0C0C0} 83.5}} & {\color[HTML]{C0C0C0} 97.5} & {\color[HTML]{C0C0C0} 82.9} \\
& LP& & & & 64.2& \multicolumn{1}{c|}{18.5}& 69.0& \multicolumn{1}{c|}{36.5}& 98.6& 89.8 & 95.3& \multicolumn{1}{c|}{73.5}& 93.9& 56.8\\ 
& &\checkmark & & & 65.1& \multicolumn{1}{c|}{20.0}& 71.2& \multicolumn{1}{c|}{33.4}& 98.9& 90.0 & 95.8& \multicolumn{1}{c|}{81.0}& 94.8& 65.9\\
& & &\checkmark & & 66.5& \multicolumn{1}{c|}{21.2}& 75.9& \multicolumn{1}{c|}{32.9}& 99.4& 90.6 & 96.7& \multicolumn{1}{c|}{82.3}& 95.2& 70.3\\
& & & &\checkmark & 93.5& \multicolumn{1}{c|}{77.0}& 97.0& \multicolumn{1}{c|}{83.7}& 99.7& 91.9 & 97.0& \multicolumn{1}{c|}{82.5}& 96.1& 78.3\\
\multirow{-6}{*}{\begin{tabular}[c]{@{}c@{}}ImageNet\\ (ViT-B)\end{tabular}}& \multirow{-4}{*}{Ours}&\checkmark &\checkmark &\checkmark & \textbf{94.0} & \multicolumn{1}{c|}{\textbf{77.6}} & \textbf{97.3} & \multicolumn{1}{c|}{\textbf{84.6}} & \textbf{99.8} & \textbf{92.0}& \textbf{97.5} & \multicolumn{1}{c|}{\textbf{84.2}} & \textbf{97.2} & \textbf{82.4} \\ \hline
& {\color[HTML]{C0C0C0} FT} & {\color[HTML]{C0C0C0} } & {\color[HTML]{C0C0C0} } & {\color[HTML]{C0C0C0} } & {\color[HTML]{C0C0C0} 98.7} & \multicolumn{1}{c|}{{\color[HTML]{C0C0C0} 89.7}} & {\color[HTML]{C0C0C0} 99.0} & \multicolumn{1}{c|}{{\color[HTML]{C0C0C0} 92.5}} & {\color[HTML]{C0C0C0} 99.7} & {\color[HTML]{C0C0C0} 94.9}& {\color[HTML]{C0C0C0} 97.3} & \multicolumn{1}{c|}{{\color[HTML]{C0C0C0} 84.5}} & {\color[HTML]{C0C0C0} 97.8} & {\color[HTML]{C0C0C0} 83.0} \\
& LP& & & & 68.0& \multicolumn{1}{c|}{27.4}& 75.1& \multicolumn{1}{c|}{38.7}& 99.3& 93.9 & 94.1& \multicolumn{1}{c|}{70.2}& 94.6& 69.0\\  
& &\checkmark & & & 71.0& \multicolumn{1}{c|}{30.4}& 78.4& \multicolumn{1}{c|}{40.1}& 99.4& 94.0 & 94.8& \multicolumn{1}{c|}{78.3}& 95.1& 72.1\\
& & &\checkmark & & 75.5& \multicolumn{1}{c|}{36.2}& 82.0& \multicolumn{1}{c|}{46.2}& 99.4& 94.2 & 95.2& \multicolumn{1}{c|}{80.2}& 95.8& 72.3\\
& & \multicolumn{1}{l}{}& \multicolumn{1}{l}{}&\checkmark & \multicolumn{1}{c}{92.6}& \multicolumn{1}{c|}{76.8}& \multicolumn{1}{c}{96.6}& \multicolumn{1}{c|}{83.6}& \multicolumn{1}{c}{99.6}& \multicolumn{1}{c|}{94.0}& \multicolumn{1}{c}{96.7}& \multicolumn{1}{c|}{82.8}& 96.5& 80.2\\
\multirow{-6}{*}{\begin{tabular}[c]{@{}c@{}}Pathoduet\\ (ViT-B)\end{tabular}} & \multirow{-4}{*}{Ours}&\checkmark &\checkmark &\checkmark & \multicolumn{1}{c}{\textbf{94.1}} & \multicolumn{1}{c|}{\textbf{77.6}} & \multicolumn{1}{c}{\textbf{97.3}} & \multicolumn{1}{c|}{\textbf{86.9}} & \multicolumn{1}{c}{\textbf{99.8}} & \multicolumn{1}{c|}{\textbf{94.9}} & \multicolumn{1}{c}{\textbf{97.6}} & \multicolumn{1}{c|}{\textbf{84.8}} & \textbf{97.5} & \textbf{82.6} \\ \hline
\end{tabular}}
\end{center}
\end{table*}

\textbf{Effectiveness of PathoTune.} The results of different foundation models using mixed combinations of prompts (Table~\ref{tab:dataset_comparison}) yield key insights: (1) Our method significantly surpasses linear probing in all scenarios, closely rivals full finetuning with just 5.9\% of the trainable parameters (More details in supplementary material). Furthermore, employing multi-modal prompts greatly enhances performance compared to single-modal usage. (2) The performance of a useful downstream adaptation method (\textit{e.g.}, our PathoTune) based on the natural visual foundation model far exceeds that of a poor downstream adaptation approach (\textit{e.g.}, linear probing) based on pathological foundation model, suggesting that efficient downstream adaptation is even more important than pretraining a pathological foundation model. (3) Transferring from a pathological foundation model shows slightly better results than a visual foundation model under the same backbone scale, indicating the optimal strategy originates from a pathological foundation model paired with an effective finetuning approach. (4) PathoTune can significantly enhance underperforming foundation models, \textit{e.g.}, ImageNet (ViT-S), elevating their performance to rival that of specialized models, \textit{e.g.}, Pathoduet (ViT-B).

\begin{figure*}[t]
  \centering
  \includegraphics[width=1.0\linewidth]{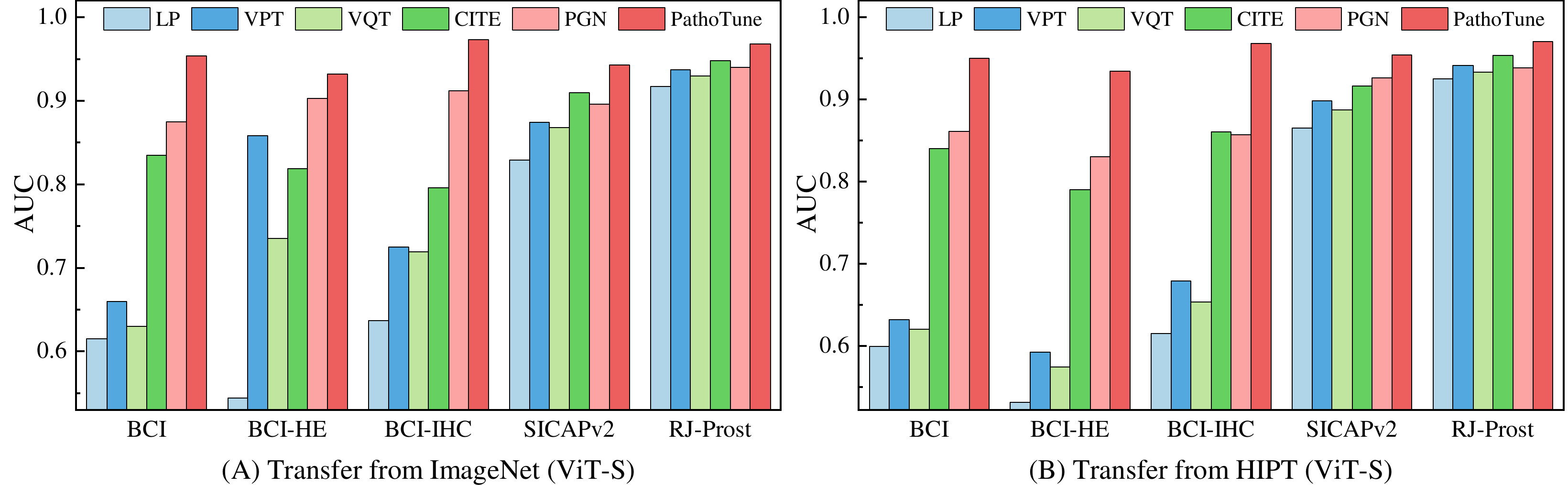}
  \caption{Comparisons of the PathoTune with other SOTA methods of PEFT.}
  \label{Fig:sota_compare}
\end{figure*}

\textbf{Comparisons with SOTAs.} We evaluate PathoTune's performance against SOTA methods including VPT~\cite{jia2022visual}, VQT~\cite{tu2023visual}, CITE~\cite{zhang2023text}, and PGN~\cite{loedeman2022prompt} across various datasets, as depicted in Fig.~\ref{Fig:sota_compare}. PathoTune consistently surpasses all compared methods, regardless of whether it transfers from the visual or pathological foundation model. The results demonstrate the superiority of PathoTune's multi-modal prompts elaborated for domain gaps over these approaches using single-modal prompts.

\begin{figure*}[t]
  \centering
  \includegraphics[width=1.0\linewidth]{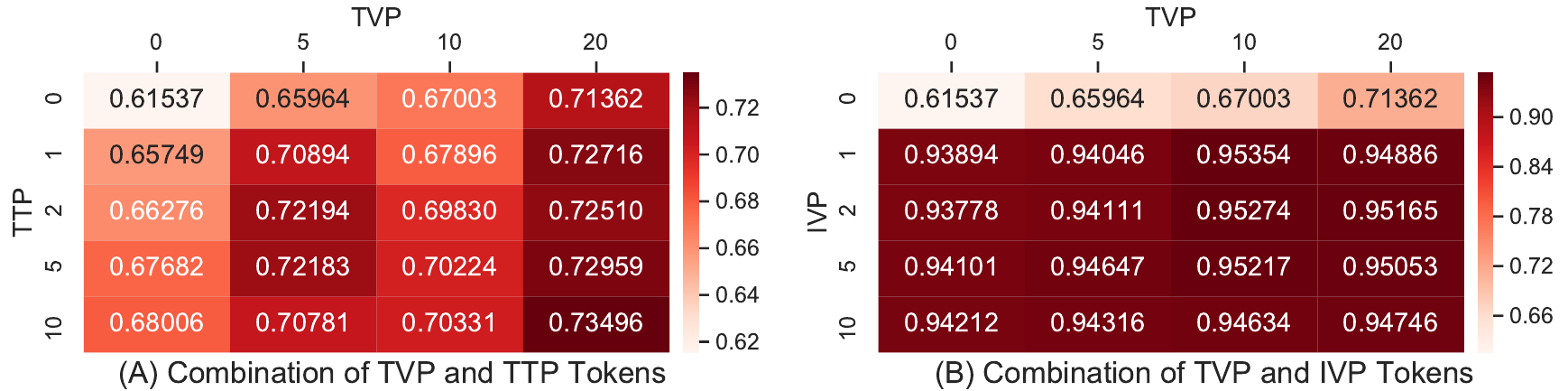}
  \caption{Comparisons of the PathoTune with different prompt combination.}
  \label{Fig:prompt_combine}
\end{figure*}

\textbf{Impacts of prompt combination.} 
We evaluate the performance of PathoTune under different combinations of prompts adapted from ImageNet (ViT-S) on the BCI dataset, with IVP and TTP taken as 0 respectively. As shown in Fig.~\ref{Fig:prompt_combine}, using a combination of prompts yields better results than using a single prompt, and IVP emerges as the most effective one.

\section{Conclusion}
In this paper, we present \textbf{PathoTune}, an innovative framework designed to adapt generalist foundation models to specialized pathological tasks through multi-modal prompt tuning. By addressing the Foundation-Task Gap and the Task-Instance Gap, we propose the Task-specific Visual Prompts, Task-specific Textual Prompts, and Instance-specific Visual Prompts. PathoTune not only surpasses SOTA methods but also remarkably outperforms pretrained pathological foundation models using linear probing, providing a new paradigm for computational pathology applications in the pretrain-finetune era.

\begin{credits}
\subsubsection{\ackname} This study was supported by Shanghai Artificial Intelligence Laboratory.

\subsubsection{\discintname}
The authors have no competing interests to declare that are
relevant to the content of this article.
\end{credits}

%
%
%
%

\bibliographystyle{splncs04}
\bibliography{egbib}





\end{document}